%% file: main.tex
\documentclass[runningheads]{llncs}
\titlerunning{Voxel2Hemodynamics}
\usepackage{graphicx}
\usepackage{subfig} 
\usepackage{makecell}
\usepackage{marvosym}
\usepackage{multirow}
\usepackage{booktabs}
\usepackage{amsmath}

\begin{document}

\captionsetup[figure]{name={Fig.},labelsep=period}
\captionsetup[table]{name={Table},labelsep=period}

\title{Voxel2Hemodynamics: An End-to-end Deep Learning Method for Predicting Coronary Artery Hemodynamics}
% Xiaoyu Yang ${ }^{1,2}$, Lijian Xu ${ }^{2,3}$ (四), Simon Yu ${ }^4$, Qing Xia ${ }^5$, Hongsheng $\mathbf{L i}^6 \&$ Shaoting Zhang ${ }^2$
% ${ }^1$ College of Electronics and Information Engineering, Tongji University
% ${ }^2$ Shanghai Artificial Intelligence Laboratory
% ${ }^3$ Centre for Perceptual and Interactive Intelligence, the Chinese University of Hong Kong
% ${ }^4$ Department of Imaging and Interventional Radiology, the Chinese University of Hong Kong
% ${ }^5$ SenseTime Research
% ${ }^6$ Department of Electronic Engineering, the Chinese University of Hong Kong
\authorrunning{${}$}
\author{Ziyu Ni ${ }^{1 (\dagger)}$, Linda Wei ${ }^{2 (\dagger)}$ , Lijian Xu ${ }^{1,3}\textsuperscript{(\Letter)}$, Simon Yu ${ }^{3}$, Qing Xia ${ }^{4}$, Hongsheng Li ${ }^{5}$ and Shaoting Zhang ${ }^{1}$}

\institute{${ }^1$ Shanghai Artificial Intelligence Laboratory\\
${ }^2$ Shanghai Jiao Tong University\\
${ }^3$ Centre for Perceptual and Interactive Intelligence, the Chinese University of Hong Kong\\
${ }^4$ SenseTime Research \\
${ }^5$ Department of Electronic Engineering, the Chinese University of Hong Kong
}
\maketitle              % typeset the header of the contribution
\begin{abstract}
Local hemodynamic forces play an important role in determining the functional significance of coronary arterial stenosis and understanding the mechanism of coronary disease progression. Computational fluid dynamics (CFD) have been widely performed to simulate hemodynamics non-invasively from coronary computed tomography angiography (CCTA) images. However, accurate computational analysis is still limited by the complex construction of patient-specific modeling and time-consuming computation. In this work, we proposed an end-to-end deep learning framework, which could predict the coronary artery hemodynamics from CCTA images. The model was trained on the hemodynamic data obtained from 3D simulations of synthetic and real datasets. Extensive experiments demonstrated that the predicted hemdynamic distributions by our method agreed well with the CFD-derived results. Quantitatively, the proposed method has the capability of predicting the fractional flow reserve with an average error of 0.5\% and 2.5\% for the synthetic dataset and real dataset, respectively. Particularly, our method achieved much better accuracy for the real dataset compared to PointNet++ with the point cloud input. This study demonstrates the feasibility and great potential of our end-to-end deep learning method as a fast and accurate approach for hemodynamic analysis.

\end{abstract}

\input{Parts/Introduction}

\input{Parts/Methods}

\input{Parts/Experiments}

\input{Parts/Conclusion}


\bibliographystyle{splncs04}
\bibliography{Parts/ref}
\end{document}

%% file: Parts/Introduction.tex
\section{Introduction}
Coronary artery disease (CAD) is one of the most common types of cardiovascular disease in the world, which is mainly caused by plaque buildup in the arterial wall \cite{RF1}. In clinical procedure, revascularization is routinely performed in the treatment of severe myocardial ischemia, where the degree of stenosis is usually regarded as the criterion for surgical intervention \cite{RF3,RF2}.
Nevertheless, biomechanical and hemodynamic alterations have been speculated to play an essential role in the pathogenesis of CAD and long-term outcomes of treatments \cite{RF5,RF4}. For instance, fractional flow reserve (FFR) has been established as the golden standard for diagnosis of intermediate stenosis in patients with chronic CAD \cite{RF6}. Besides, wall shear stress (WSS) is a measure of the shear force exerted on the arterial inner surface and the abnormal WSS has been found to exert negative influence of endothelial function \cite{RF7}. In this context, the quantitative evaluation of hemodynamic characteristics would contribute to the early diagnosis of coronary diseases. Computational fluid dynamics (CFD) based methods have been extensively examined to obtain hemodynamic parameters non-invasively \cite{RF9,RF8,RF10}. Using the patient-specific geometry and boundary conditions obtained from medical imaging data, in vivo hemodynamics could be reproduced accurately and non-invasively. The fidelity of a CFD model in reproducing hemodynamics relies on the accurate patient-specific modeling and various assumptions involved in model setup \cite{RF8}. Another important limitation of CFD modeling is associated with the vast computational resources and long computing times required. In order to promote the clinical application of hemodynamics, it is necessary to develop a novel method in balance of the accuracy and computational cost.

Deep learning has achieved state-of-the-art results in automated segmentation of coronary computed tomography angiography (CCTA) images \cite{RF11,RF12} and emerged as a potential approach to improve the efficiency of traditional physical modeling methods \cite{RF15,RF14,RF13,RF16}. Advanced deep learning algorithms and high-performance GPUs could greatly reduce computing times while ensuring high accuracy. Several studies have already been introduced concerning predictions of coronary artery hemodynamics from point cloud \cite{RF13} or geometrical features simply \cite{RF15,RF14,RF16}, which however are limited due to the ignorance of patient-specific geometrical and physiological features. Itu et al. put forward a DNN-based model to predict FFR of synthetically generated coronary anatomies with extracted feature input \cite{RF14}. Li et al. simply fed point cloud to the model to give prediction on hemodynamics of non-ideal cardiovascular model \cite{RF13}, but the patient-specified information has not been taken into consideration. Numerical studies have demonstrated that such simplified models could lead to marked deviations of model outputs from real in vivo hemodynamic conditions, which highlights the importance of patient-specifically modeling \cite{RF18,RF17}. The hemodynamic of the coronary artery tree is complicated by its intricate geometry with massive coronary branches, upstream hemodynamics as well as downstream resistance ratio dominated by the micro circulation. Therefore, accurate prediction of the coronary hemodynamics required the patient-specific modeling, which should be carefully considered in deep learning method.

In this work, an end-to-end deep learning method has been developed to predict the hemodynamic parameters (e.g. velocity, pressure, WSS and FFR) with the input of CCTA images and physiological parameters. Both synthetic and real dataset of represented stenosis were generated based on the patient-specific geometry and imported into the CFD pipeline to produce the training data. The model was subsequently trained with those generated hemodynamic data under different physiological conditions. We further compared the hemodynamic distributions predicted by the deep learning model and CFD.

%% file: Parts/Methods.tex
\begin{figure}[h]
    \centering
    \includegraphics[width=\textwidth]{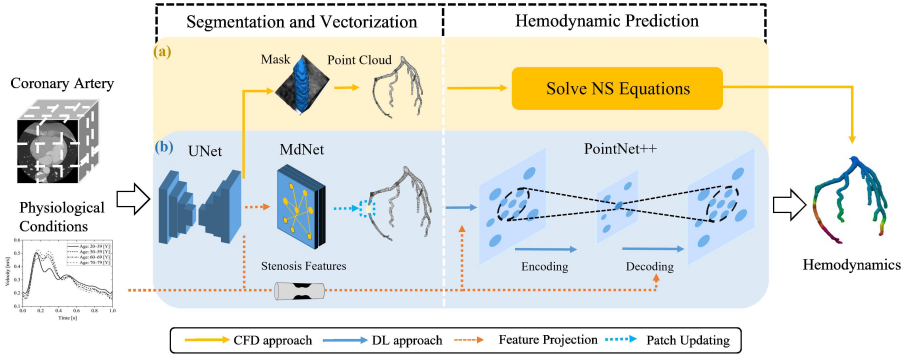}
    \caption{Setup of  conventional computational hemodynamic modeling (a) and our proposed end-to-end deep learning approach (b).}
    \label{fig:my_label}
\end{figure}
\section{Proposed Method}
Our framework included three parts: 1) 3D UNet \cite{3dunet} for coronary artery segmentation, and 2) Mesh deformation network (MdNet) \cite{ICLR} for point cloud construction, and 3) PointNet++ \cite{pointnet++} for target prediction. The image features were extracted from CCTA images by UNet and subsequently fused into MdNet and PointNet++. During training, the three models were jointly optimized to achieve higher performance. 
\subsubsection{Segmentation and Vectorization:}
Firstly, 3D UNet was utilized to segment the inner diameter of coronary arteries and extract the geometric features from CCTA images (e.g. stenosis-related features shown in Fig. 1). High-level perceptual features from the last three convolutional layers of UNet decoder were projected and concatenated as auxiliary input for MdNet and PointNet++. To guide the segmentation process, we used cross-entropy loss ($L_{ce}$) and dice loss ($L_{dice}$), as shown in equation (1):
\begin{equation}
\min _{\theta_u} \sum_{i=1}^n(L_{c e}\left(f_u\left(x_{u_i} ; \theta_u\right), y_{u_i}\right)+L_{\text {dice }} (f_u\left(x_{u_i} ; \theta_u\right), y_{u_i} ))
\end{equation}
where $f_u$ represents 3D UNet and $\theta_u$ represents its paramters. $n$ is the number of batch size and $x_{u i}$ is the input patch with 32x32x32 cropped size.
Secondly, the intermediate point cloud was constructed in a coarse-to-fine way by MdNet, which was composed of three graph convolutional network \cite{gcn} blocks with graph unpooling layer. Following Pixel2Mesh \cite{pixel2mesh}, mesh loss was applied to constrain the shape  during mesh deformation, as shown in equation (2):
\begin{equation}
\min _{\theta_m} \sum_{i=1}^n L_{\text {mesh }}\left(f_m\left(x_m, f_{u-d} ; \theta_m\right), y_{m_i}\right)
\end{equation}
where $f_m$ represents MdNet and $\theta_m$ represents its parameters. $x_m$ is the initial ellipsoid mesh input with 162 vertices and $f_{u-d}$ is the extracted features from 3D UNet decoder. $y_{m i}$ is the deformed mesh with 2562 vertices.
% \begin{figure}[h]
%     \centering
%     \includegraphics[width=\textwidth,height=0.7\textwidth]{figures/fig2_2.eps}
%     \caption{The ground truth (CFD) and model predicted hemodynamics of blood flow through an idealized LAD stenosis with a reference diameter of 0.3cm and a 50\% diameter stenosis. Rest and hyperemic myocardial blood flow was respectively simulated by adjusting the downstream microcirculatory resistance. The velocity is set at zero along the luminal boundary. Abbreviations: LAD, left anterior descending artery.}
%     \label{fig:my_label}
% \end{figure}
\subsubsection{Hemodynamic Prediction:}
The whole point cloud of coronary tree was merged from the patch data and fed into PointNet++. During training and testing phase, input points were sampled to a fixed size (e.g. 20000 points) for our STENOSIS-900 dataset. To promote the capability of our model at stenosis region, the projected high-level features were extracted from UNet and jointly guide the decoding process to make more accurate prediction. Mean absolute error (mae) was employed as the loss function ($L_{mae}$) for training PointNet++, as shown in equation (3):
\begin{equation}
\min _{\theta p} \sum_{i=1}^n L_{\text {mae }}\left(f_p \left(x_{p_i}, f _{u-d}, h_t ; \theta_p\right), y_{p_i}\right)
\end{equation}
where $f_p$ represents PointNet++ and $\theta_p$ represents its parameters. $h_t$ is the extra physiological input and $y_{p i}$ is hemodynamic parameters calculated by CFD.

\subsubsection{Loss and Training Strategy:}
The overall loss function for our proposed framework is shown in equation (4):
\begin{equation}
    \lambda_1L_{ce} +\lambda_2L_{dice}+\lambda_3L_{mesh}+\lambda_4L_{mae} 
\end{equation}
where $\lambda_1,\lambda_2,\lambda_3,\lambda_4$ are hyper parameters to adjust the weight between losses.
To optimize the framework from end to end, whole point cloud was constructed by the jointly pretrained UNet and MdNet and subsquently used as initial input for PointNet++. In training phase, whole point cloud was updated by patch-result at each iteration, and all three models were jointly optimized. Normalized mean absolute error (NMAE) was used as evaluation metric, as shown in equation (5):
\begin{equation}
\mathrm{NMAE}=\frac{1}{n} \frac{\sum_{i=1}^n\left|p_i-\hat{p}_i\right|}{\operatorname{max}|p|-\operatorname{min}|p|}
\end{equation}
where $p_i$ and $\hat{p_i}$ represent CFD and our method result respectively. Adam optimizer was utilized with initial learning rate of $10^{-3}$ to train the framework. Our method was implemented on pytorch and trained with 4 NVIDIA GeForce GTX 1080 Ti GPUs. For real dataset, the end-to-end inference time was less than 20 seconds with one GPU.

% \begin{figure}[h]
%     \centering
%     \includegraphics[width=\textwidth,height=\textwidth]{figures/fig3_4.eps}
%     \caption{FFR Distributions of coronary artery tree in three representative examples utilizing CFD (a),  our model (b) and PointNet++ (c). The specific FFR values were recorded at the 2cm downstream of stenoses locations (marked by red circles). Abbreviations: LAD, left anterior descending artery; LCx, left circumflex artery; RCA, right coronary artery.}
%     \label{fig:my_label}
% \end{figure}
\begin{figure}[h]
    \centering
    \includegraphics[width=\textwidth,height=0.7\textwidth]{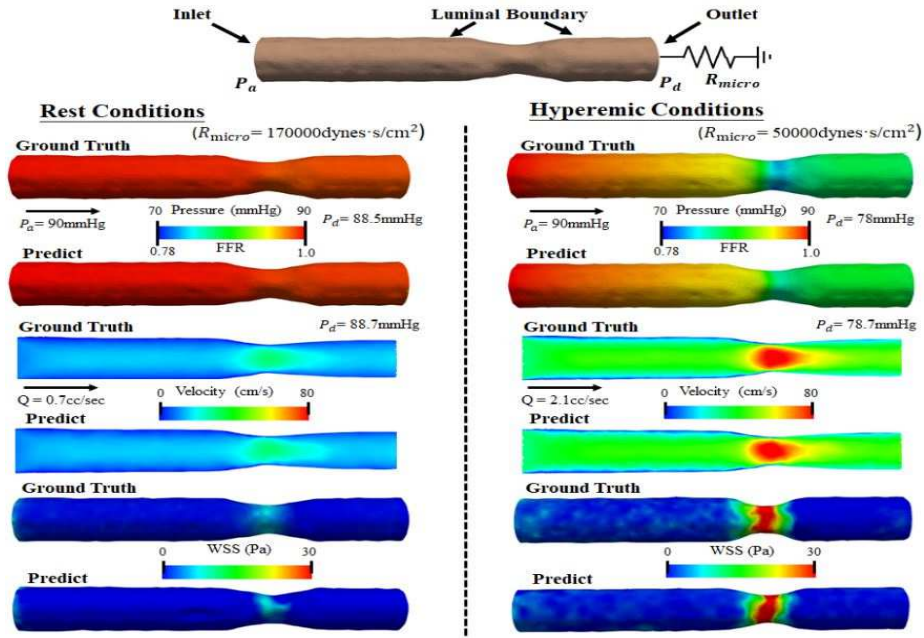}
    \caption{The ground truth (CFD) and model predicted hemodynamics of blood flow through an idealized LAD model with a reference diameter of 0.3 cm and a 50\% diameter stenosis. Rest and hyperemic myocardial blood flow was respectively simulated by adjusting the downstream microcirculatory resistance. The velocity is set at zero along the luminal boundary. }
    \label{fig:my_label}
\end{figure}

%% file: Parts/Experiments.tex
\begin{figure}[h]
    \centering
    \includegraphics[width=\textwidth,height=0.88\textwidth]{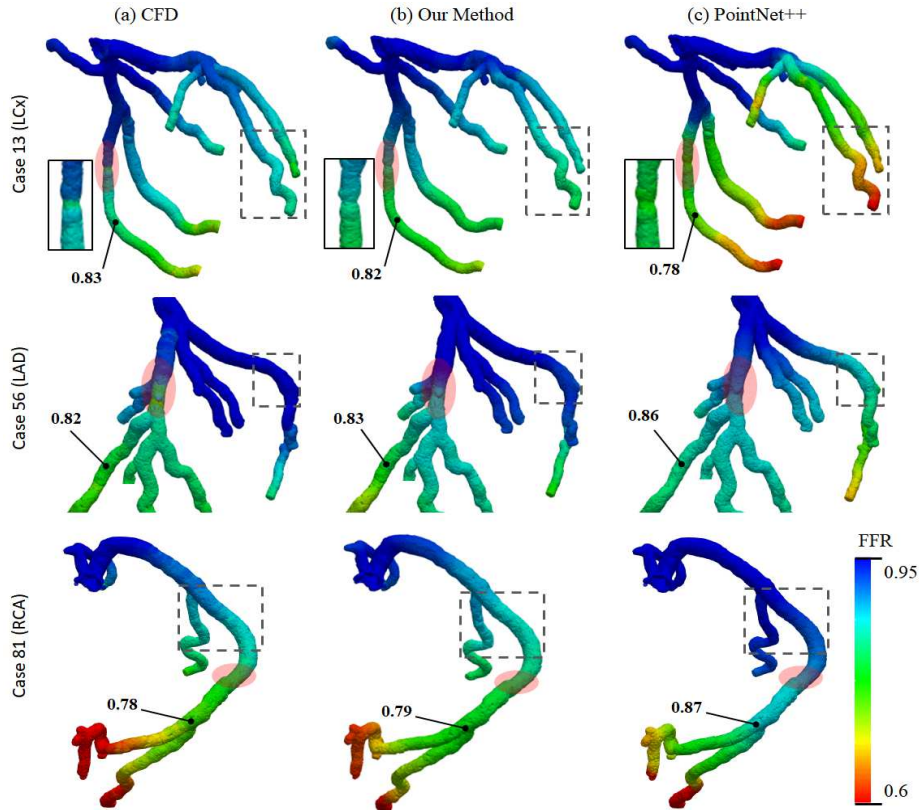}
    \caption{FFR distributions of coronary artery tree in three representative examples utilizing CFD (a),  our model (b) and PointNet++ (c). The specific FFR values were recorded at the 2 cm downstream of stenosis locations marked by red circles. The differences were marked by dash boxes.}
    \label{fig:my_label}
\end{figure}
%Abbreviations: LAD, left anterior descending artery; LCx, left circumflex artery; RCA, right coronary artery.

\section{Experiments}
\subsection{Datasets}
\subsubsection{Synthetic and Real Dataset:}
The idealized LAD vessel models were generated with a unified length of 2 cm, a reference diameter of 0.3 cm and a single stenosis. The location of stenosis beginning varied from 0.5 cm to 1.5 cm, with the length of stenosis region ranging from 0.2 cm to 0.4 cm. The degree of stenosis ($DS =\left(1-r_{s t e n}/r\right) \times 100 \%$) varied from 50\% to 70\%, 
% $DS =\left(1-r_{s t e n}/r_{L A D}\right) \cdot 100 \%$
% \begin{equation}
% DS =\left(1-\frac{r_{s t e n}}{r_{L A D}}\right) \cdot 100 \%
% \end{equation}
where $r_{s t e n}$ is the minimum radius in the stenosis region, and  $r$ is the unified radius of the idealized vessel.
A total of 150 CCTA images were further collected from one clinical institution retrospectively. The average in-plane resolution and slice thickness is 0.038 cm and 0.047 cm, respectively. The stenoses were randomly generated at three main branches (i.e. left anterior descending artery(LAD), left circumflex artery(LCx), and right coronary artery(RCA)) and sampled between zero and two. A total number of 900 CCTA images and masks were produced and named STENOSIS-900, where 720 cases generated from the original 120 data were used for training, and 180 cases generated from the other original 30 data were used for testing.

\subsubsection{Hemodynamic Dataset:}
The hemodynamic dataset was produced by the automated pipeline (shown in Fig. 1a) of imaged-based CFD simulation  consisting of the following steps: 1) CCTA images were firstly segmented to reconstruct geometrical models of coronary arteries. Subsequently, the fluid domain of the geometrical model was divided using tetrahedral elements, followed by a mesh refinement with prism layers. Mesh sensitivity studies were further conducted and verified that the adopted mesh density was sufficient to yield numerically acceptable results; 2) All simulations were herein conducted with OpenFOAM package where the Navier-Stokes (NS) equations were discretized and solved with finite volume method. Gauss upwind was employed for the spatial discretization. Coronary artery flow was assumed to be an incompressible fluid with a density of $1060 \rm kg/m^3$ governed by the unsteady three-dimensional NS equations. The typical Carreau model was employed to calculate the blood viscosity. Additionally, arterial walls were assumed to be rigid where the non-slip boundary conditions were imposed. Resistance outflow boundary was applied by placing a resistance distal to each outlet to simulate the physiological flow division \cite{RF10}. 
% \begin{figure}[htbp]
%     \centering
%     \includegraphics[width=\textwidth]{figures/fig2.eps}
%     \caption{The ground truth (CFD) and model predicted hemodynamic results of blood flow through an idealized LAD stenosis with a reference diameter of 0.3cm and a 65\% diameter reduction stenosis. Rest and hyperemic myocardial blood flow was respectively simulated by adjusting the downstream microcirculatory resistance. The velocity is set at zero along the luminal boundary.Abbreviations: LAD, left anterior descending artery.}
%     \label{fig:my_label}
% \end{figure}
\subsection{Results}
 An idealized LAD model with a 50\% diameter stenosis was herein employed to evaluate the model performance at rest and hyperemic conditions (see Fig. 2). Hyperemic conditions were assumed to assess model predictions under myocardial stress similar to the environment of clinical measurement. The constant pressure ($P_a$ = 90 mmHg) was applied at the inlet boundary, while the downstream microcirculatory resistance ($R_{micro} = 170000$ or $50000$ dynes·s/$\rm cm^2$) was prescribed at the outlet boundary to simulate the rest and hyperemic conditions, respectively. Compared with the CFD results, our model predicted consistent distributions of all hemodynamic metrics (e.g. FFR, pressure drop, velocity and WSS) at rest condition. Besides, the hemodynamic alterations (e.g. larger pressure drop/velocity/WSS across the stenosis region and smaller downstream FFRs) were accurately predicted when switching rest condition to hyperemic mycardial blood flow conditions. FFR was a pressure-basd metric, which was calculated as ratio of mean blood pressure at the corresponding location and the mean aortic pressure under hyperemic condition. Quantitatively, our model achieved downstream FFR of 0.874 at hyperemia conditions, while CFD derived value was 0.867. 

Fig. 3 shows the CFD and model predicted FFR results of three patient-specific cases, where the stenosis is located at the three branches (i.e. LAD, LCx, RCA) respectively. Our predicted FFR distributions agreed well with the CFD results throughout the coronary tree for all three cases. Besides, the specific model-derived FFR value downstream the stenosis was quite close to the CFD derived result. Clinicians usually adopted a fixed cut-off value (e.g. FFR < 0.8) to identify patients with myocardial ischemia. With the proposed method, our model successfully identified the non-ischemic patients (case 13 and case 56) and ischemic patient (case 81) although all three patients suffered moderate coronary stenosis. RCA stenosis was responsible for the myocardial ischaemia based on the predicted FFRs for case 81, which agreed with the clinical descriptions collected. 

\begin{table}[h]
\caption{FFR results of idealized LAD and STENOSIS-900 datasets at vessel level. NMAE is regarded as the evaluation metric.}
\centering
\begin{tabular}{cccccc}
\toprule
\multirow{2}{*}{Method} & \multirow{2}{*}{ Idealized LAD } & \multicolumn{4}{c}{ STENOSIS-900 } \\
 & & All branches & LAD & LCx & RCA \\
\midrule \makecell[c]{PointNet++} & \makecell[c]{$0.007 \pm 0.006$} & $0.034 \pm 0.020$ & $0.036 \pm 0.020$ & $0.033 \pm 0.022$ & $0.031 \pm 0.021$ \\
 \makecell[c]{Ours} & \makecell[c]{$0.005 \pm 0.004$} & $0.025 \pm 0.022$ & $0.027 \pm 0.019$ & $0.025 \pm 0.018$ & $0.021 \pm 0.020$ \\
\bottomrule
\end{tabular}
\end{table}

% \begin{tabular}{|c|c|c|c|c|c|}
% \hline \multirow{2}{*}{ Method } & \multirow{2}{*}{$\begin{array}{c}\text { Idealized } \\
% \text { LAD }\end{array}$} & \multicolumn{4}{|c|}{ STENOSIS-900 } \\
% \hline & & All branches & $\mathrm{LAD}$ & $\mathrm{LCx}$ & $\mathrm{RCA}$ \\
% \hline PointNet++ & $0.007 \pm 0.006$ & $0.034 \pm 0.020$ & $p .036 \pm 0.020$ & $0.033 \pm 0.022$ & $0.031 \pm 0.021$ \\
% \hline Ours & $0.005 \pm 0.004$ & $0.025 \pm 0.022$ & $0.024 \pm 0.019$ & $0.023 \pm 0.018$ & $0.019 \pm 0.020$ \\
% \hline
% \end{tabular}

\noindent To further examine the effectiveness of the proposed method, PointNet++ was trained for ablation study. With the point cloud of three-dimensional spatial coordinates, PointNet++ was capable of predicting the general distribution of FFR but performed poorly at the distal narrow branches and bifurcation of coronary artery (highlighted by dashed boxes in Fig. 3c), which may attribute to the absence of stenosis features. As shown in Table 1, NMAE was 0.005 ± 0.004 and 0.025 ± 0.022 for the idealized and real dataset respectively. It further verified quantitatively that our method improved the prediction accuracy compared to PointNet++. Moreover, a significant correlation was observed between the estimations of CFD-derived FFRs and their counterparts predicted with our method$ (\textit{r} = 0.758, \textit{p} \textless 0.001)$, with the FFR varied in a small range around a mean value of 0.021 as shown in the Bland–Altman plot (see Fig. 4).

%\begin{table}[htbp]
% \caption{FFR results of idealized LAD and STENOSIS-900 dataset at vessel level. NMAE is regarded as the evaluation metric.}
% \centering 

% \begin{tabular}{|c|c|cccc|}
% \hline 
% \makecell[c]{Method} & { Idealized LAD } & \multicolumn{4}{|c|}{ STENOSIS-900 } \\
% \cline { 3 - 6 } & & All branches & LAD & LCx & RCA \\
% \hline \makecell[c]{PointNet++} & \makecell[c]{$0.007 \pm 0.006$} & $0.034 \pm 0.020$ & $0.036 \pm 0.020$ & $0.033 \pm 0.022$ & $0.031 \pm 0.021$ \\
% \hline \makecell[c]{Ours} & \makecell[c]{$0.005 \pm 0.004$} & $0.025 \pm 0.022$ & $0.027 \pm 0.019$ & $0.025 \pm 0.018$ & $0.021 \pm 0.020$ \\
% \hline
% \end{tabular}
% \begin{tabular}{|c|cc|c|c|c|}
% \hline  Proposed Method  & Idealized & \multicolumn{4}{|c|}{ STENOSIS-900 } \\
% \cline { 3 - 6 } & LAD & All branches & LAD & LCx & RCA \\
% \hline PointNet++ & $0.007 \pm 0.006$ & $0.034 \pm 0.020$ & $0.036 \pm 0.020$ & $0.033 \pm 0.022$ & $0.031 \pm 0.021$ \\
% \hline Ours & $0.005 \pm 0.004$ & $0.025 \pm 0.022$ & $0.024 \pm 0.019$ & $0.023 \pm 0.018$ & $0.019 \pm 0.020$ \\
% \hline
% \end{tabular}
% \end{table}

\begin{figure*} [t!]
	\centering
	\subfloat[\label{fig:a}]{
		\includegraphics[width=0.4\linewidth]{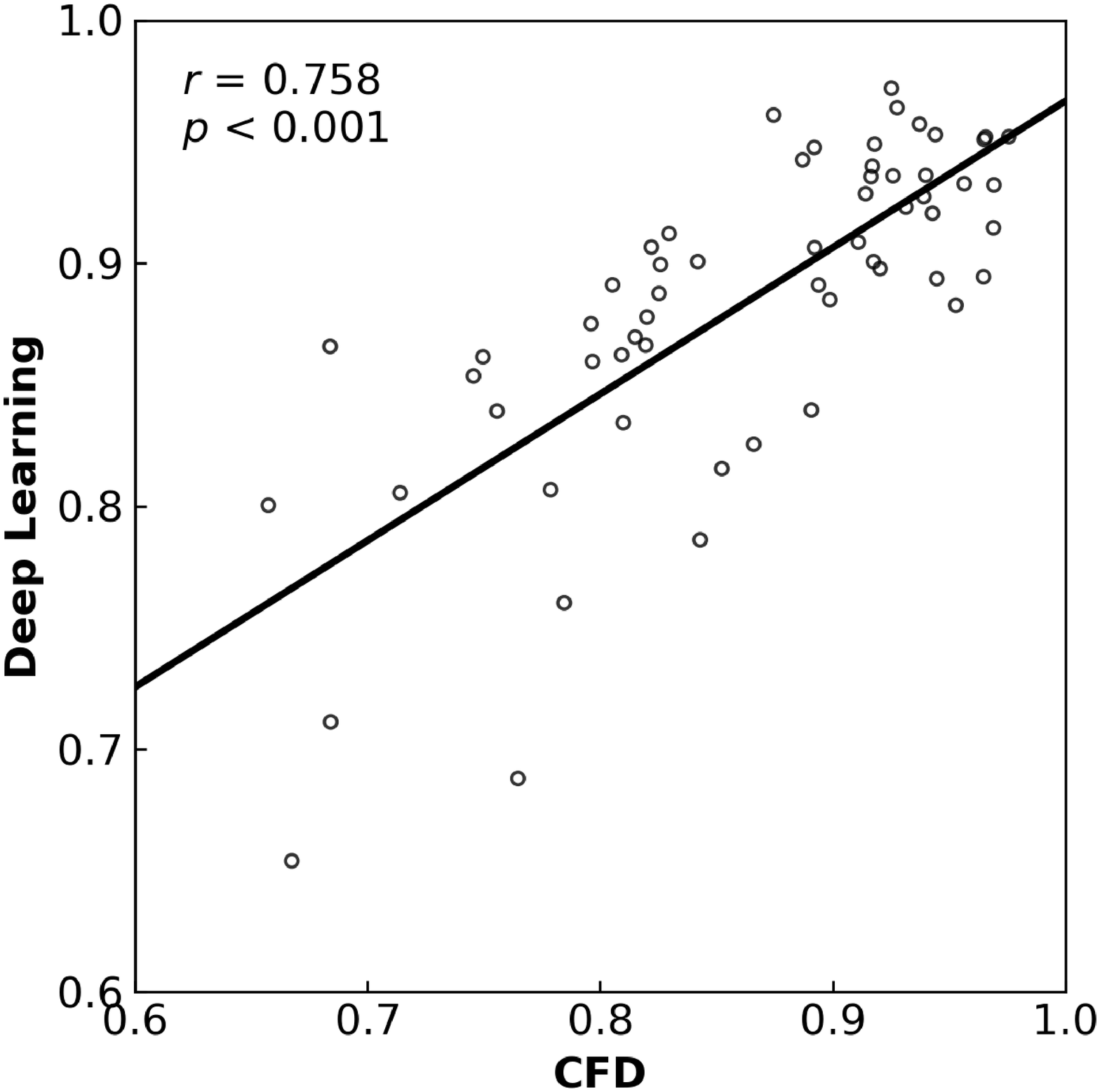}}
	\subfloat[\label{fig:b}]{
		\includegraphics[height=0.4\textwidth]{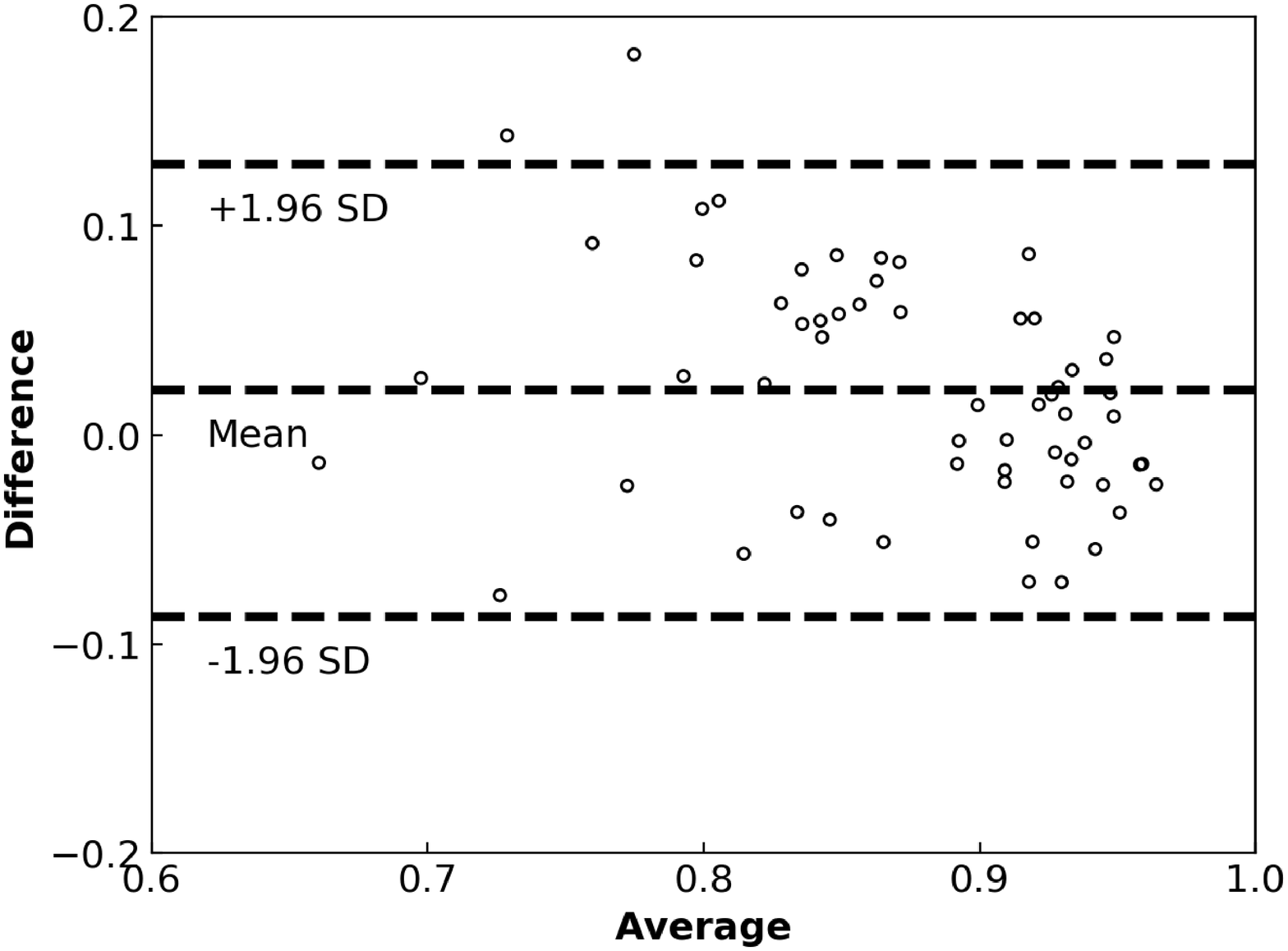}}
	\caption{Correlation (a) and Bland–Altman (b) plots of FFRs predicted by CFD vs. Deep Learning. FFRs were recorded at locations about 2cm downstream from stenoses. All results are  from STENOSIS-900 test set.}
	\label{fig:my_label} 
\end{figure*}

%% file: Parts/Conclusion.tex
\section{Conclusion}
In this work, we presented an end-to-end deep learning approach to predict coronary artery hemodynamic indicators (e.g. velocity, pressure, WSS and FFR). By incorporating image features and physiological parameters, our method demonstrated capability to predict hemodynamics under various physiological conditions and yielded more accurate results compared to the conventional PointNet++. Our method could be generalized to estimate the hemodynamic metrics of other vascular related diseases, which may contribute to identifying patients at high risk of cardiovascular disease and patient-specific treatment in future.